\documentclass[%
aps,
 reprint,
 amsmath,amssymb,
prper,
floatfix
]{revtex4-2}

\usepackage{rotating}
\usepackage{booktabs}

\newcommand{\sattr}[1]{{\it #1}}

\newcommand{\scibank}{{\sc SciEntsBank}}
\newcommand{\beetle}{{\sc Beetle}}

\begin{document}

\title[Automated Short Answer Grading with a Large Language Model]{Performance of the Pre-Trained Large Language Model GPT-4\\on Automated Short Answer Grading}



\begin{abstract}
Automated Short Answer Grading (ASAG) has been an active area of machine-learning research for over a decade. It promises to let educators grade and give feedback on free-form responses in large-enrollment courses in spite of limited availability of human graders. Over the years, carefully trained models have achieved increasingly higher levels of performance.  More recently, pre-trained  Large Language Models (LLMs) emerged as a commodity, and an intriguing question is how a general-purpose tool without additional training compares to specialized models. We studied the performance of GPT-4 on the standard benchmark 2-way and 3-way datasets SciEntsBank and Beetle, where in addition to the standard task of grading the alignment of the student answer with a reference answer, we also investigated withholding the reference answer. We found that overall, the performance of the pre-trained general-purpose GPT-4 LLM is comparable to hand-engineered models, but worse than pre-trained LLMs that had specialized training.
\end{abstract}

\keywords{Automated Short Answer Grading, Large Language Model, SciEntsBank, Beetle, GPT}

\author{Gerd Kortemeyer}
 \email{kgerd@ethz.ch}
 \affiliation{%
 Educational Development and Technology, ETH Zurich, 8092 Zurich, Switzerland
}

\date{\today}

\maketitle

\section{Introduction}
Providing meaningful feedback to learners is one of the most important tasks of instructors~\cite{bransford2000people}. yet it can also become one of the most work-intensive or even tedious tasks. Particularly for large-enrollment courses, lack of grading personnel can limit this feedback to automatically gradable closed-answer formats such as multiple-choice or numerical inputs. This limitation might be overcome by using Artificial Intelligence (AI) solutions~\cite{seo2021};  it is therefore not surprising that when it comes to the use of AI in higher education, assessment and evaluation are the most prominent topics~\cite{crompton2023}, and acceptance of this technology for education is increasing based on its perceived usefulness~\cite{zhang2023}. In particular, studies on Automated Short Answer Grading (ASAG)~\cite{burrows2015,haller2022survey} are highly relevant for educators to extend the limits of what can be assessed at large scales.

It is impossible to do justice to the spectrum of sophisticated ASAG methods in this short study; Burrows, Gurevych, and Stein provide an excellent overview up to 2015~\cite{burrows2015}; Haller, Aldea, Seifert, and Strisciuglio look at later developments up to 2022~\cite{haller2022survey}. The latter survey notes a particular shift in recent years as models are moving from hand-engineered features to representation-learning approaches, which draw their initial training data from large text corpora~\cite{haller2022survey} (``pre-trained''). However, most models used for ASAG still have in common that they are explicitly trained or fine-tuned for particular grading tasks, and datasets used in competitions such as SemEval~\cite{dzikovska2013} thus include training and testing items. By contrast, recently publicly released Large Language Models (LLMs) such as GPT-4~\cite{gpt4} and Bard~\cite{bard} have not only been pre-trained from large text corpora, but subsequently extensively fine-tuned following general instead of task-specific strategies.  Their users are neither expected nor actually able to further train or fine-tune the model, and an intriguing question is how these out-of-the-box general-purpose tools perform compared to specially trained or fine-tuned models.

In this study, GPT-4 is prompted to grade the items from two standard datasets, \scibank{} and \beetle~\cite{dzikovska2013}, which allows comparison of precision, recall, and F1-score (or weighted F1-score in case of 3-way items) to legacy and state-of-the-art ASAG models. \scibank{} covers general science questions for 3rd to 6th grade, while \beetle{} covers questions and student answers from a tutorial system for basic electricity and electronics.

The standard judgment method is to compare the student answer to a reference answer, but in addition, it was also investigated if GPT-4 can adequately grade the student answers based on the question alone. For the latter task, the model would need to draw on its own pre-training from its text corpus to independently judge the correctness of the student answer.

\section{Methodology}
The \scibank{} and \beetle{} datasets~\cite{dzikovska2013} were downloaded from kaggle~\cite{dataset}. They included both training and test data. The training data were discarded, while the test data included the $504$~items and $14,186$~student answers and their reference grading that were used for this study. As no training took place, the distinction between unseen answers (UA), unseen questions (UQ), and unseen domains (UD) that the dataset provided was dropped for this study, since all items were ``unseen.''

Each item in the datasets contains a question, a reference answer, and student answers including their reference grade. The items came in two versions:
\begin{itemize}
\item a 2-way version, where each student answer is either \sattr{correct} if it is complete and correct paraphrase of the reference answer or \sattr{incorrect} otherwise, and 
\item a 3-way version, where an additional judgment of \sattr{contradictory} replaces some of the \sattr{incorrect} labels if the student answer explicitly contradicts the reference answer.
\end{itemize}

The XML-coded items were translated into prompts for the GPT-4 API, see Figure~\ref{fig:prompts} for an example. Each item was graded with and without providing a reference answer. The definitions of the judgment criteria for grading were taken from SemEval-2013~\cite{dzikovska2013}.

\begin{figure*}
\begin{center}
\begin{minipage}{0.96\textwidth}
\small
\begin{verbatim}
<?xml version="1.0"?>
<question testSet="unseen-domains" id="WA_52b" module="WA">
  <questionText>Johnny drove to the store with his father one cold and rainy night.
    They had only driven a short distance when the windows "fogged up" on the inside.
    What was it about the windows that caused the "fog" to form on them?</questionText>
  <referenceAnswers>
    <referenceAnswer id="WA_52b-a1">The windows were cooler than the water vapor in the air,
    causing the vapor to condense.</referenceAnswer>
  </referenceAnswers>
  <studentAnswers>
    <studentAnswer id="WA.52b.171.1" accuracy="correct">The windows in the car were cold
    because it was cold outside. There was lots of water vapor in the car, so it stuck to
    the cold windows and changed into condensation water.</studentAnswer>
    <studentAnswer id="WA.52b.174.1" accuracy="correct">They were cold and since Johnny
    probably turned the heat ...
    \end{verbatim}
\end{minipage}

\begin{tabular}{p{0.48\textwidth}|p{0.48\textwidth}}
\begin{center}
{\bf With Reference Answer}
\end{center}&
\begin{center}
{\bf Without Reference Answer}
\end{center}
\\
\small
{\bf Role:} You are an assistant grading short student answers. You provide your grades solely in a CSV table with the columns "ID" and "correctness, where you list the full (un-shortened) ID and your grading result.".
You grade based on a reference answer that will be provided.
You grade as "correct" if the student answer is a complete and correct paraphrase of the reference answer.
You grade as "contradictory" if the student answer explicitly contradicts the reference answer.
You grade as "incorrect" otherwise.
&
\small
{\bf Role:} You are an assistant grading short student answers. You provide your grades solely in a CSV table with the columns "ID" and "correctness, where you list the full (un-shortened) ID and your grading result.".
You grade as "correct" if the student answer is correct and comprehensive.
You grade as "incorrect" otherwise.
\\
\small
{\bf Content:} Grade the student answers to the question
"Johnny drove to the store with his father one cold and rainy night. They had only driven a short distance when the windows "fogged up" on the inside. What was it about the windows that caused the "fog" to form on them?".
The reference answer is given as
"The windows were cooler than the water vapor in the air, causing the vapor to condense.".
The student answers are listed below in the format "ID:answer".

WA.52b.171.1:The windows in the car were cold because it was cold outside. There was lots of water vapor in the car, so it stuck to the cold windows and changed into condensation water.

WA.52b.174.1:They were cold and since Johnny probably turned the heat \ldots
&
\small
{\bf Content:} Grade the student answers to the question
"Johnny drove to the store with his father one cold and rainy night. They had only driven a short distance when the windows "fogged up" on the inside. What was it about the windows that caused the "fog" to form on them?".
The student answers are listed below in the format "ID:answer".

WA.52b.171.1:The windows in the car were cold because it was cold outside. There was lots of water vapor in the car, so it stuck to the cold windows and changed into condensation water.

WA.52b.174.1:They were cold and since Johnny probably turned the heat \ldots
\end{tabular}
\end{center}

\caption{Original XML-code of a 3-way item and the generated prompts for its evaluation with and without providing a reference answer.}
\label{fig:prompts}
\end{figure*}

For $6$~of the $504$~items, errors occurred during evaluation, which led to $58$~of the $28,372$~student statements receiving no or invalid grades. The invalid grades were \sattr{unclear}, \sattr{creative}, \sattr{epoch}, \sattr{accurate}, and \sattr{correc} [sic]. These missing or invalid student grades were counted as neither positives nor negatives. 

Subsequently, the precision, recall, and (weighted) F1-score were calculated:
\begin{description}
\item[Precision:] Out of all the \sattr{correct} grades given by a model, how many were actually correct?
?\item[Recall (or Sensitivity):] Out of all the actual correct student answers, how many were graded as \sattr{correct}?
\item[F1-score:] Harmonic mean of precision and recall;  a way to balance the trade-off between precision and recall.
\end{description}
In the 3-way scenario, the above characteristics are correspondingly calculated for the classes \sattr{contradictory} and \sattr{incorrect}, and a weighted average is calculated for these class F1-scores to form the weighted F1-score (w-F1).

\section{Results}
\subsection{Precision, Recall, and F1-Scores}
Table~\ref{tab:results} shows the precision, recall, and F1-scores for \scibank{} and \beetle{} for the 2-way and 3-way items, as well as for the scenario where the reference answer was withheld. For the 3-way scenario, the individual-class results and the weighted F1-score (w-F1) are provided.

\begin{sidewaystable}
\caption{Results for precision, recall, and F1-scores for \scibank{} and \beetle{} in the 2-way, 3-way, and no-reference-answer scenarios. For the 3-way scenario, the individual-class results and the weighted F1-score (w-F1; also referred to as micro-averaged F1-score) are provided.}\label{tab:results}
\begin{tabular*}{\textheight}{@{\extracolsep\fill}lcccccccccccccccc}
\toprule
&	\multicolumn{3}{c}{2-way}&	\multicolumn{10}{c}{3-way}&\multicolumn{3}{c}{No Reference Answer}\\\cmidrule(lr){2-4}\cmidrule(lr){5-14}\cmidrule(lr){15-17}
&			&	     &  &\multicolumn{3}{c}{\sattr{correct}}	&\multicolumn{3}{c}{\sattr{contradictory}}	&\multicolumn{3}{c}{\sattr{incorrect}}	\\\cmidrule(lr){5-7}\cmidrule(lr){8-10}\cmidrule(lr){11-13}			
&	Prec.&	Rec.&	F1&	Prec.&	Rec.	&F1&	Prec.	&Rec.&	F1	&Prec.&	Rec.	&F1&	w-F1	 &Prec.&	Rec.	&$F1$
\vspace*{2mm}\\
{\sc SciEnts-}&\\
{\sc Bank}&	0.788	&0.705	&0.744&	0.717&	0.825&	0.767&	0.696&	0.581&	0.633&	0.754	&0.679	&0.715&	0.729	&0.697	&0.768	&0.731\\

\beetle&	0.657	&0.572	&0.611&	0.635	&0.672&	0.653&	0.680&	0.199	&0.308&	0.426&	0.672&	0.522&	0.516&	0.581&	0.739&	0.651\\
\botrule
\end{tabular*}
\end{sidewaystable}
Looking at the precision and recall, an outlier is the recall on \sattr{contradictory} in the 3-way \beetle{} dataset: a large number of student answers that were labelled as \sattr{contradictory} were not recognized as such, but simply as \sattr{incorrect} (as evidenced by the low precision on \sattr{incorrect}).

GPT-4 generally performs better on \scibank{} than on \beetle{}. For \scibank, the model showed its highest performance on the 2-way task (F1=0.744), followed closely by the no-reference scenario (F1=0.731), with the 3-way scenario in last place (w-F1=0.729). For \beetle, the no-reference scenario showed the highest performance (F1=0.651), followed by the 2-way (F1=0.611) and 3-way (w-F1=0.516) scenarios. In other words, for \beetle{}, providing a reference answer lowered its performance on correctly judging the student answers.

\subsection{Comparison to Specialized ASAG Models}
Table~\ref{tab:compare} shows a comparison of specifically trained models versus the out-of-the-box GPT-4 model. At the time of the SemEval-2013 competition~\cite{dzikovska2013}, had GPT-4 been around, it would have won the competition for 3-way \scibank, and it would have outperformed all but one competing models in the unseen questions (UQ) category. In these specifically trained models, performance strongly depends on what was ``seen'' and what was ``unseen.''

Newer systems perform better, in particular those of the BERT~\cite{devlin2018} LLM family. These models are pre-trained and then specifically trained for   \scibank{} and \beetle{} using for example PyTorch~\cite{paszke2019}.
Unfortunately, for the highly successful  roberta-large model~\cite{poulton2021}, the performance was not separately reported for the different  `unseen'' categories, and no 3-way grading was performed.

\begin{sidewaystable}
\caption{Comparison of (weighted) F1-scores for different ASAG systems and GPT-4.}\label{tab:compare}
\begin{tabular*}{\textheight}{@{\extracolsep\fill}llcccccccccccc}
\toprule
&&\multicolumn{7}{c}{\scibank}&\multicolumn{5}{c}{\beetle}\\
\cmidrule(lr){3-9}\cmidrule(lr){10-14}
&&\multicolumn{3}{c}{2-way}&\multicolumn{3}{c}{3-way}&No. Ref.&\multicolumn{2}{c}{2-way}&\multicolumn{2}{c}{3-way}&No. Ref.\\
\cmidrule(lr){3-5}\cmidrule(lr){6-8}\cmidrule(lr){10-11}\cmidrule(lr){12-13}
Model&Year&UA&UQ&UD&UA&UQ&UD&&UA&UQ&UA&UQ&
\vspace*{2mm}\\
CoMeT~\cite{dzikovska2013}&2013&0.77&0.58&0.67&0.71&0.52&0.55&&0.83&0.70&0.73&0.49\\
SoftCardinality~\cite{dzikovska2013}&2013&0.72&0.74&0.71&0.65&0.63&0.62&&0.77&0.64&0.62&0.45\\
Sultan et al.~\cite{sultan2016,saha2018}&2016&0.69&0.70&0.71&0.57&0.62&0.60\\
Saha et al.~\cite{saha2018}&2018&0.79&0.70&0.72&0.71&0.64&0.61\\
GCN-DA~\cite{tan2023}&2020&&&0.73&&&063\\
SFRN+~\cite{li2021}&2021&0.78&0.64&0.67&0.65\footnote{\label{micro}not stated if macro-averaged F1 or weighted (micro-averaged) F1 was reported}&0.49\footref{micro}&0.47\footref{micro}&&0.89&0.70&0.67\footref{micro}&0.55\footref{micro}\\
BERT~\cite{filighera2022}&2022&&&&0.73&0.60&0.62&&&&0.71&0.57\\
roberta-large~\cite{poulton2021}&2021&\multicolumn{3}{c}{--- 0.81\footnote{\label{nou}the model was specifically trained, but no separate information on UA, UQ, and UD was provided} ---}&\multicolumn{3}{c}{}&&\multicolumn{2}{c}{--- 0.91\footref{nou} ---}\\
GPT-4 &2023&\multicolumn{3}{c}{--- 0.74 ---}&\multicolumn{3}{c}{--- 0.73 ---}& 0.73&\multicolumn{2}{c}{--- 0.61 ---}&\multicolumn{2}{c}{--- 0.52 ----}&0.65\\
\botrule
\end{tabular*}
\end{sidewaystable}
Overall, the performance of the pre-trained general-purpose GPT-4 LLM is comparable to hand-engineered models, but worse than pre-trained LLMs that had specialized training. 
\section{Limitations}
Since GPT is a probabilistic model, running it again, possibly at a different temperature, is likely going to yield different results. However, due to the already large amount of computing required for one run, and in light of the high statistics gained from over $500$~items, only one run was considered here. Also, different prompts from the ones shown in Fig.~\ref{fig:prompts} may result in higher or lower performance.

OpenAI, the company behind GPT, does not release information about what constituted the text corpus used for training. Though unlikely, since the datasets are only available as ZIP-files and in XML-format, there is still a possibility that \scibank{} and \beetle{} had been used for training. When asked about \scibank{}, though, the model stated that it was not familiar with a dataset or source by that name; GPT-4 performed better on \scibank{} than on \beetle{}, for which it  stated that it is a known dataset in the domain of natural language processing and educational research. The model, however, demonstrated ignorance when asked about any specific details regarding Johnny, his father, and the windows in the scenario quoted in Fig.~\ref{fig:prompts}, making it unlikely that it had seen the text before.

\section{Discussion}
The last five years saw the strong emergence of Deep-Learning-based models for ASAG. These models generally exhibit higher performance than hand-engineered models, but still strongly depend on training, which may be pre-training or task-specific. LLMs usually come pre-trained, but the extend of that pre-training greatly varies: while details on GPT-4's text corpus are proprietary, it can be assumed that it was trained and fine-tuned with orders of magnitude more data than for example BERT~\cite{devlin2018}. However, as this study shows, the difference in pre-training can be more than made up by the BERT-family's openness to additional task-specific training by the user.

At least for the grade-school educational content covered by the datasets in this study, GPT-4 performs ASAG at a performance level comparable to hand-engineered systems from five years ago. It does so even without the need for providing reference answers. There are strong indications that this ability would extend to higher education, for example university-level physics content~\cite{kortemeyer2023could}, and that automated grading of open-ended assessment content is possible beyond short answers~\cite{kortemeyer2023can}. In addition, a general-purpose LLM can give more tailored feedback than simple \sattr{correct}/\sattr{incorrect} judgments, which has high potential for learning from short answer grading~\cite{jordan2009}.

A problem with general-purpose tools like GPT-4~\cite{gpt4} and Bard~\cite{bard} is that they are running in the cloud. When it comes to grade-relevant student data, the question of data security and privacy cannot be ignored, which may limit the applicability of this approach to ASAG. An alternative for a model that might also not need additional training, but which could be locally installed, is Llama~2~\cite{llama}, However, preliminary studies by the author indicate that Llama~2 does generally not perform as well as GPT-4.

\section{Conclusion}
The performance of the general-purpose Large Language Model GPT-4 on Automated Short Answer Grading does not reach that of specifically trained Deep-Learning models, but it is comparable to that of earlier hand-engineered ASAG models. A clear advantage of GPT-4 is that it does not need to be specifically trained for the task and can be used ``out-of-the-box,'' which has the potential to turn it into a commodity for educators. In addition to not needing additional training, GPT-4 can also perform ASAG without the need for providing reference answers, at least at the grade-school level covered by the datasets used in this study and likely at the introductory higher-education level.

\section*{Acknowledgements}
The author would like to thank Julia Chatain for her help in connecting to the GPT API.
\section*{Declarations}
\subsection*{Availability of data and material}
The benchmark datasets \scibank{} and \beetle~\cite{dzikovska2013} are available from kaggle~\cite{dataset}. Code and calculated data are made available as supplemental material alongside this paper from \url{https://www.polybox.ethz.ch/index.php/s/mByv0od7uscm3VV} (the file readme.txt in the downloadable package explains the code and data files) .

\subsection*{Funding}
Not applicable.

\bibliographystyle{unsrt}
\bibliography{shortanswer}


\end{document}